\documentclass{svproc}
\usepackage{url}
\usepackage{multirow}
\usepackage{listings}
\usepackage{algorithm}
\usepackage{algorithmic}
\usepackage{siunitx,booktabs}

\begin{document}
\mainmatter              % start of a contribution
\title{Exploiting Tournament Selection for Efficient Parallel Genetic Programming}
\titlerunning{Exploiting Tournament Selection for Efficient Parallel Genetic Programming}  % abbreviated title (for running head)
\author{Darren M. Chitty}%\inst{1}}
\authorrunning{Darren M. Chitty} % abbreviated author list (for running head)
\tocauthor{Darren M. Chitty}

\institute{Department of Computer Science,\\ University of Bristol, Merchant Venturers Bldg,\\ Woodland Road, BRISTOL BS8 1UB\\ \email{darrenchitty@googlemail.com}}  \maketitle              % typeset the title of the contribution

\begin{abstract}
Genetic Programming (GP) is a computationally intensive technique
which is naturally parallel in nature. Consequently, many attempts
have been made to improve its run-time from exploiting highly
parallel hardware such as GPUs. However, a second methodology of
improving the speed of GP is through efficiency techniques such as
subtree caching. However achieving parallel performance and
efficiency is a difficult task. This paper will demonstrate an
efficiency saving for GP compatible with the harnessing of
parallel CPU hardware by exploiting tournament selection.
Significant efficiency savings are demonstrated whilst retaining
the capability of a high performance parallel implementation of
GP. Indeed, a 74\% improvement in the speed of GP is achieved with
a peak rate of 96 billion GPop/s for classification type problems.
\keywords{Genetic Programming, HPC, Computational Efficiency}
\end{abstract}

\vspace{-0.7cm}
\section{Introduction}\vspace{-0.2cm}
\label{sec:Introduction} Genetic Programming (GP) \cite{Koza:1992}
is widely known as being highly computationally intensive. This is
due to candidate GP programs being typically evaluated using an
interpreter which is an inefficient method of running a program
due to the use of a conditional statement at each step in order to
ascertain which instruction to execute. Moreover, GP is a
population based technique with generally large population sizes
of candidate GP programs. As such, there have been many
improvements to the execution speed of GP from a compiled approach
or direct machine code, exploiting parallel computational hardware
or finding efficiencies within the technique such as caching
common subtrees. Recent advances in computational hardware have
led to multi-core architectures such that very fast parallel
implementations of GP have been implemented but a drawback is that
although they are fast, they are not efficient. Techniques such as
subtree caching are difficult to use with these parallel
implementations. Searching for common subtrees can incur a
significant time cost slowing execution speeds.

This paper introduces an efficiency saving that can be made by
exploiting the characteristics of tournament selection which can
be easily implemented within a CPU based parallel GP approach.
Consequently, a considerable performance gain can be made in the
execution speed of GP. The paper is laid out as follows: Section
\ref{sec:Background} describes GP and prior methods of improving
the speed and efficiency of the technique. Section
\ref{sec:EfficientTournSelection} introduces a tournament
selection strategy embedded with a highly parallel GP model and
demonstrates efficiency savings with regards classification tasks.
Finally, Section \ref{sec:nonmod} demonstrates significant
enhancements by consideration of evaluated solutions surviving
intact between generations.

\vspace{-0.2cm}
\section{Background}\label{sec:Background}\vspace{-0.2cm}
Genetic Programming (GP) \cite{Koza:1992} solves problems by
constructing programs using the principles of evolution. A
population of candidate GP programs are maintained and evaluated
as to their respective effectiveness against a target objective.
New populations are generated using the genetic operators of
selection, crossover and mutation. Selection is usually conducted
with tournament selection where a subset of GP programs compete to
be selected as parents based on their fitness. The evaluation of a
GP program is typically achieved by \emph{interpreting} it against
a set of fitness cases. GP is computationally intensive as a
result of using an interpreter, maintaining typically large
populations of programs and often, using a large volume of fitness
cases such as for classification tasks.

Recently, with the advent of multi-core CPUs and many core GPUs,
the focus has been on creating highly parallel implementations of
GP achieving speedups of several hundred
\cite{Cano:2012,Chitty:2012,Augusto:2013,Chitty:2016,Chitty:2017}.
However, prior to the move to parallel architectures the primary
method of improving the speed of GP was through efficiency
savings. A simple methodology is to reduce the number of fitness
cases by dynamic sampling of the more difficult instances
\cite{Gathercole:1994}. A further selection strategy known as
Limited Error Fitness (LEF) was investigated by Gathercole and
Ross for classification problems whereby an upper bound on the
number of permissible misclassifications was used to terminate
evaluations \cite{Gathercole:1997}. Maxwell implemented a time
rationing approach where each GP program was evaluated for a fixed
time \cite{Maxwell:1994}. Tournament selection was then performed
and the fittest candidate GP program was declared the winner.
Smaller programs could evaluate more fitness cases. Teller used a
similar technique known as the \emph{anytime} approach
\cite{Teller:1994}.

With regards tournament selection Teller and Andre introduced a
technique known as the Rational Allocation of Trials (RAT)
\cite{Teller:1997} whereby population members are evaluated on a
small subset of individuals. A prediction model is then used to
test if further evaluations should be made to establish the
\emph{winners} of tournaments with a 36x speedup but only small
populations are considered and regression type problems. Park et
al. implemented a methodology whereby the fitness of the best
candidate GP program within the current population is tracked and
when evaluating a GP program, if the accumulated fitness becomes
worse than this best found then the evaluation is terminated and
the fitness approximated \cite{Park:2013}. Speedups of four fold
were reported with little degradation in overall fitness. Finally,
Poli and Langdon ran GP backwards by generating tournaments for
the whole GP process and working backwards detecting offspring
that are not sampled by tournament selection and not evaluating
them or their parents if all offspring are not sampled
\cite{Poli:2006}. However, most of these methods do not accurately
reflect tournament winners thereby changing the evolutionary path.

\vspace{-0.2cm}
\section{Improved Efficiency Through Tournament Selection}\label{sec:EfficientTournSelection}\vspace{-0.2cm}
The typical methodology of GP is to completely evaluate a
population of GP programs to ascertain their fitness. A new
population is then constructed using the GP operators of
selection, crossover and mutation. The selection process typically
chooses two \emph{parents} from the current population to generate
two offspring programs using crossover and mutation. The most
widely used selection operator within GP is tournament selection
whereby $t$ GP programs are randomly selected from the current
population to form a \emph{tournament}. The program within this
\emph{tournament} with the best fitness is then selected as a
parent.

However, tournament selection can be exploited by considering an
alternative evaluation methodology. Rather than evaluating each
candidate GP program over every single fitness case before moving
onto the next candidate, consider the opposite approach whereby
all programs are evaluated upon a single fitness case before
considering the next fitness case. Using this approach, the
fitness levels of candidate GP programs slowly build up and
facilitates comparisons between programs during the evaluation
process. Consequently, if a set of tournaments is generated prior
to the evaluation stage it is possible to ascertain which
candidate GP programs within a given tournament reach a stage
whereby they cannot possibly \emph{win}. Moreover, if a candidate
GP program is deemed unable to \emph{win} any of the tournaments
it is involved in before all the fitness cases have been evaluated
upon, there is clearly no reason to continue to evaluate it.
Consequently an efficiency saving can be realised using this
approach. This technique can be described as \emph{smart sampling}
whereby only the minimum fitness cases necessary to establish
\emph{losers} of tournaments is required.
\\\\
\textbf{Example:} Consider a classification problem whereby the
fitness metric is the sum of correct classifications with a
tournament size of two to select a potential parent. Using ten
fitness cases the first GP program of the tournament correctly
classifies the first six cases and the second incorrectly
classifies them. Effectively the second GP program in the
tournament cannot possibly \emph{win} as the maximum correct
classifications it can now achieve is four from the remaining
fitness cases. If this second GP program is not involved in any
other tournament then there is no value in continuing to evaluate
it for the remaining fitness cases as it will definitely not be
selected as a parent. Consequently, a 40\% efficiency saving can
be achieved on the evaluation of the second GP program within the
tournament.
\\\\
To realise this potential efficiency saving a new GP evaluation
model is proposed whereby at each generation a set of tournaments
composed of randomly selected candidate GP programs from the
population are generated prior to the evaluation. Using a system
whereby two selected programs generate two offspring, this set
consists of population size $n$ tournaments each of which contain
tournament size $t$ randomly selected programs. Each fitness case
is then evaluated by every member of the population before the
next fitness case is considered. Before a program is evaluated on
the given fitness case it is checked to establish if it is has
effectively \emph{lost} all of the tournaments it is involved in
and if so labelled as a \emph{loser} and not to be further
evaluated. The efficient tournament selection work suggested here
ensures the same candidate GP programs \emph{win} tournaments as
with standard GP. Algorithm \ref{alg:EfficientTourn} provides a
high level overview of the efficient tournament selection model.

\vspace{-0.4cm}

\begin{algorithm}[]
\caption{Efficient Tournament Selection}
\label{alg:EfficientTourn}
\begin{algorithmic}[1]
    \STATE{initialise population}
    \FOR{number of generations}
        \STATE{generate set of tournaments for generation of next
        population}
        \FOR{each fitness case}
            \FOR{each population member}
                \IF{population member has not lost all
                its tournaments}
                    \STATE{evaluate population member on given
                    fitness case and update fitness}
                \ENDIF
            \ENDFOR
        \ENDFOR
        \STATE{generate new population using tournament winners}
    \ENDFOR
\end{algorithmic}
\end{algorithm}

\vspace{-1.2cm}

\begin{algorithm}[]
\footnotesize \caption{Tournament Check} \label{alg:TournCheck}
\begin{algorithmic}[1]
    \FOR{each tournament given population member is involved in}
        \STATE{identify current best fitness of population
        members in given tournament}
        \IF{fitness of given population member plus the potential
        additional fitness from the remaining fitness cases is lower
        than this best fitness}
            \STATE{given population member has already \emph{lost} this tournament}
        \ENDIF
    \ENDFOR
    \IF{given population member cannot possibly \emph{win} any of its tournaments}
        \STATE{stop evaluating population member}
    \ENDIF
\end{algorithmic}
\end{algorithm}

\vspace{-0.4cm}

To determine if a candidate GP program has \emph{lost} all of the
tournaments it is involved in, each tournament needs to be checked
once the current fitness case has been evaluated upon by all
candidate GP programs in the population. Thus, each candidate GP
program maintains the subset of tournaments it is involved in. For
each of these tournaments, the currently best performing candidate
GP program in the tournament is identified and its fitness
compared with the fitness of the candidate GP program under
consideration. If using a given fitness metric it can be
ascertained that the GP program under consideration cannot beat
this best then it is designated as having \emph{lost} this
tournament. If a program is deemed to have \emph{lost} all of the
tournaments it is involved in then it is designated as requiring
no further evaluation. A high level description of tournament
checking is shown in Algorithm \ref{alg:TournCheck}. Using this
approach all the candidate GP programs that would \emph{win} the
tournaments using a standard implementation of GP will still
\emph{win} their respective tournaments. Consequently, there is no
disturbance to the evolutionary path taken by GP but an efficiency
saving can be achieved.

\subsection{Efficient Tournament Selection and a Fast
GP Approach}\vspace{-0.1cm}

Given that the goal of efficient tournament selection is to reduce
the computational cost of GP and hence improve the speed, it is
only natural that the technique should be able to operate within a
fast parallel GP model. Integrating the efficient tournament
selection model with a GPU implementation would prove difficult
without compromising the speed of the approach as communication
across GPU cores evaluating differing GP programs is difficult.
The alternative platform is a CPU based parallel GP
\cite{Chitty:2012} which introduced a two dimensional stack
approach to parallel GP demonstrating significantly improved
execution times. A multi-core CPU with limited parallelism was
used with the two dimensional stack model to exploit the cache
memory and reduce interpreter overhead. In fact, this model
actually operates similarly to efficient tournament selection, GP
programs are evaluated in parallel over blocks of fitness cases.
Once all the candidate GP programs have been evaluated on a block
of fitness cases, the next block of fitness cases is considered.
Using blocks of fitness cases provided the best utilisation of
cache memory and hence the best speed.

Subsequently, the efficient tournament selection model is
implemented within this two dimensional stack GP model using a CPU
and instead of evaluating candidate GP programs on a single
fitness case at a time, they are evaluated on a larger block of
fitness cases. Once the block of fitness cases has been evaluated
upon then candidate GP programs that may have \emph{lost} all of
their respective tournaments can be identified. A block of 2400
fitness cases is used which was identified by Chitty
\cite{Chitty:2012} as extracting the best performance from the
cache memory and efficiency from reduced reinterpretation of
candidate GP programs.

\vspace{-0.2cm}
\subsection{Initial Results}
\vspace{-0.1cm} In order to evaluate if efficient tournament
selection can provide a computational saving it will be tested
against three classification type problems. The first two are the
Shuttle and KDDcup classification problems available from the
Machine Learning Repository \cite{machinelearning:2010} consisting
of 58,000 and 494,021 fitness cases respectively. The GP function
set for these problems consists of {*, /, +, -, $>$, $<$, ==, AND,
OR, IF} and the terminal values the input features or a constant
value between -20,000.0 and 20,000.0. The third problem is the
Boolean 20-multiplexer problem \cite{Koza:1992} with the goal to
establish a rule which takes address bits and data bits and
correctly outputs the value of the data bit which the address bits
specify. The function set consists of {AND,OR,NAND,NOR} and the
terminal set consists of {A0-A3, D0-D15}. There are 1048576
fitness cases which can be reduced using bit level parallelism
such that each bit of a 32 bit variable represents a differing
fitness case reducing fitness cases to 32768.

The results for these experiments were generated using an i7 2600
Intel processor running at 3.4GHz with four processor cores each
able to run two threads of execution independently. The algorithms
used were compiled using Microsoft Visual C++. Table
\ref{tab:params} provides the GP parameters that were used
throughout the work presented in this paper. Each experiment was
averaged over 25 runs for a range of differing tournament sizes to
demonstrate how efficiency can change dependant on the selection
pressure.

\vspace{-0.2cm}

\begin{table}[!ht]
\footnotesize \centering \caption{GP parameters used throughout
results presented in this paper\vspace{-0.2cm}}
\begin{tabular}{rc|rc}
\toprule
Population Size : & 4000 & Maximum Generations : & 50\\
Maximum Tree Depth : & 50 & Maximum Tree Size : & 1000\\
Probability of Crossover : & 0.50 & Probability of Mutation : & 0.50\\
\bottomrule
\end{tabular} \centering
\label{tab:params}
\end{table}

\vspace{-0.6cm}

In order to use efficient tournament selection a fitness metric is
required which establishes if a given candidate GP program cannot
\emph{win} a tournament. This metric is described as comparing the
classification rates of the best performing GP program in a
tournament with the rate of the candidate GP program under
consideration. If the performance of the best is greater than that
of the program under consideration whilst also assuming that the
all the remaining fitness cases are correctly classified then the
program under consideration cannot possibly \emph{win}. This can
also be described as the candidate GP program under consideration
being \emph{mathematically} unable to \emph{win} the given
tournament.

Total efficiency saving is measured as the number of fitness cases
not evaluated by each GP program multiplied by their size divided
by the sum of the size of all GP programs evaluated multiplied by
the number of fitness cases. It should be noted that this work is
concerned with efficiency in the training phase of GP and not
classification accuracy. These rates are provided merely as a
demonstration that the same results are achieved between
techniques.

Table \ref{tab:firstResults} demonstrates the performance of a
standard GP approach using the 2D stack model \cite{Chitty:2012}
and the efficient tournament selection model for a range of
tournament sizes. Note that for two of the problem instances, as
the tournament size increases, the average GP tree size similarly
increases which obviously increases the execution time of GP. Also
note that there is no deviation from the classification accuracy
from both approaches as would be expected as there has been no
deviation from the evolutionary path. However, in all cases, an
efficiency saving has been observed. The greatest efficiency
saving is made with the lowest levels of tournament size as a
result of the non-sampled issue whereby some members of the
population are not involved in any tournaments \cite{Poli:2006b}.
If a candidate GP program is not involved in any tournaments then
there is no value in evaluating it. Additionally, in cases of low
selection pressure, a GP program is likely to be involved in few
tournaments and thus a poor solution can quickly \emph{lose} all
its tournaments. Also note that as the tournament size increases
to ten or greater the efficiency savings begin to improve once
more. This effect is due to an increased probability of a highly
fit GP program being in any given tournament making it easier to
identify weak solutions at an earlier stage.

\begin{center}
\begin{table*}[!h]
\scriptsize \centering \caption{Results from a comparison between
a standard implementation of GP and the efficient tournament
selection method.}
\begin{tabular}{c
                c
                S[table-number-alignment=center,separate-uncertainty,table-figures-uncertainty=1, table-figures-integer = 2, table-figures-decimal = 2]
                c
                S[table-number-alignment=center,separate-uncertainty,table-figures-uncertainty=1, table-figures-integer = 3, table-figures-decimal = 2, table-column-width=20mm]
                S[table-number-alignment=center,separate-uncertainty,table-figures-uncertainty=1, table-figures-integer = 2, table-figures-decimal = 2, table-column-width=19mm]
                S[table-number-alignment=center,separate-uncertainty,table-figures-uncertainty=1, table-figures-integer = 3, table-figures-decimal = 2, table-column-width=19mm]
                c
                }
\toprule
{\multirow{3}{1.1cm}{\centering Problem}} & {\multirow{3}{0.9cm}{\centering Tourn. Size}} & {\multirow{3}{1.4cm}{\centering Class Accuracy (\%)}} & {\multirow{3}{1.15cm}{\centering Av. Tree Size}} & {\multirow{3}{1.7cm}{\centering Standard GP Execution Time (s)}} & \multicolumn{3}{c}{Efficient Tournament Selection GP}\\
\cmidrule{6-8} & & & & & {\multirow{2}{1.6cm}{\centering Efficiency Saving (\%)}} & {\multirow{2}{1.6cm}{\centering Execution Time (s)}} & {\multirow{2}{1.0cm}{\centering Speedup}}\\
& & & & & & &\\
\midrule
\multirow{10}{1.1cm}{Shuttle} & 3 & 83.96 \pm 3.67 & 43.02 & 13.27 \pm 3.02 & 10.07 \pm 1.39 & 12.31 \pm 2.66 & 1.078x \\
& 4 & 87.92 \pm 5.34 & 46.57 & 14.04 \pm 4.49 & 8.18 \pm 1.84 & 13.22 \pm 4.05 & 1.062x \\
& 5 & 88.10 \pm 6.47 & 42.07 & 12.72 \pm 2.17 & 7.84 \pm 1.87 & 12.10 \pm 1.97 & 1.051x \\
& 6 & 89.23 \pm 6.19 & 44.49 & 13.59 \pm 3.53 & 8.37 \pm 1.62 & 12.86 \pm 3.25 & 1.057x \\
& 7 & 88.93 \pm 6.42 & 43.64 & 13.20 \pm 4.32 & 7.81 \pm 2.56 & 12.53 \pm 3.86 & 1.053x \\
& 8 & 91.44 \pm 5.99 & 41.34 & 12.64 \pm 3.14 & 8.36 \pm 2.24 & 12.00 \pm 2.81 & 1.053x \\
& 9 & 92.02 \pm 6.68 & 44.56 & 13.50 \pm 2.62 & 9.25 \pm 2.03 & 12.75 \pm 2.47 & 1.059x \\
& 10 & 90.53 \pm 6.24 & 41.98 & 12.83 \pm 3.70 & 8.44 \pm 2.32 & 12.14 \pm 3.31 & 1.057x \\
& 20 & 90.10 \pm 6.75 & 47.20 & 14.44 \pm 5.69 & 8.65 \pm 2.83 & 13.67 \pm 5.26 & 1.057x \\
& 30 & 86.43 \pm 7.36 & 47.72 & 14.39 \pm 5.69 & 8.26 \pm 2.91 & 13.73 \pm 5.34 & 1.048x \\

\midrule
\multirow{10}{1.1cm}{KDDcup} & 3 & 93.71 \pm 6.84 & 52.03 & 130.11 \pm 28.87 & 12.57 \pm 0.65 & 113.83 \pm 25.16 & 1.143x \\
& 4 & 95.96 \pm 5.37 & 54.03 & 136.38 \pm 28.21 & 11.37 \pm 1.07 & 121.61 \pm 24.41 & 1.121x \\
& 5 & 95.52 \pm 5.29 & 55.04 & 138.10 \pm 35.76 & 11.01 \pm 1.66 & 123.79 \pm 32.55 & 1.116x \\
& 6 & 95.58 \pm 5.78 & 47.65 & 120.61 \pm 25.67 & 11.50 \pm 1.52 & 107.92 \pm 23.42 & 1.118x \\
& 7 & 92.71 \pm 8.09 & 50.78 & 127.83 \pm 34.22 & 11.25 \pm 1.98 & 114.51 \pm 30.50 & 1.116x \\
& 8 & 96.26 \pm 5.24 & 51.91 & 129.11 \pm 37.06 & 11.74 \pm 1.32 & 114.97 \pm 32.58 & 1.123x \\
& 9 & 96.19 \pm 5.36 & 56.98 & 143.57 \pm 50.50 & 11.64 \pm 2.28 & 128.27 \pm 45.40 & 1.119x \\
& 10 & 95.40 \pm 6.28 & 58.16 & 149.34 \pm 63.38 & 12.42 \pm 1.58 & 132.02 \pm 55.25 & 1.131x \\
& 20 & 95.91 \pm 5.37 & 64.90 & 162.10 \pm 87.00 & 13.80 \pm 2.31 & 141.25 \pm 75.01 & 1.148x \\
& 30 & 94.64 \pm 7.12 & 65.95 & 163.46 \pm 79.64 & 12.68 \pm 2.51 & 143.81 \pm 70.15 & 1.137x \\

\midrule
\multirow{10}{1.1cm}{20-Mult.} & 3 & 61.46 \pm 0.88 & 187.28 & 53.07 \pm 7.81 & 6.46 \pm 0.11 & 49.85 \pm 7.25 & 1.065x \\
& 4 & 63.44 \pm 1.18 & 159.25 & 46.91 \pm 8.19 & 3.76 \pm 0.18 & 45.63 \pm 7.87 & 1.028x \\
& 5 & 64.85 \pm 1.35 & 154.58 & 46.46 \pm 6.87 & 2.99 \pm 0.23 & 45.62 \pm 6.69 & 1.018x \\
& 6 & 66.61 \pm 1.78 & 156.98 & 46.86 \pm 8.38 & 2.91 \pm 0.22 & 46.12 \pm 8.22 & 1.016x \\
& 7 & 67.02 \pm 1.75 & 141.64 & 42.69 \pm 6.65 & 3.04 \pm 0.28 & 41.97 \pm 6.51 & 1.017x \\
& 8 & 67.36 \pm 2.06 & 159.75 & 47.06 \pm 13.01 & 2.99 \pm 0.36 & 46.19 \pm 12.94 & 1.019x \\
& 9 & 68.38 \pm 1.74 & 132.88 & 41.41 \pm 6.98 & 3.24 \pm 0.32 & 40.60 \pm 7.03 & 1.020x \\
& 10 & 68.61 \pm 2.25 & 128.93 & 40.19 \pm 7.99 & 3.38 \pm 0.40 & 39.50 \pm 7.90 & 1.017x \\
& 20 & 71.17 \pm 2.06 & 135.81 & 41.37 \pm 8.85 & 4.36 \pm 0.46 & 40.09 \pm 8.70 & 1.032x \\
& 30 & 72.61 \pm 2.22 & 157.60 & 45.52 \pm 10.29 & 4.81 \pm 0.60 & 43.92 \pm 9.86 & 1.036x \\
\bottomrule
\end{tabular} \centering
\label{tab:firstResults}
\end{table*}
\end{center}

\vspace{-1.0cm}

The KDDcup classification problem demonstrates the greatest
efficiency savings with up to a 13.8\% saving. The multiplexer
problem demonstrates the lowest efficiency saving as a result of
the lower accuracy achieved being a more difficult problem.
Consider that as the fitness improves within the population during
the evolutionary process, identifying weak candidate GP programs
becomes easier to achieve at an earlier stage thereby increasing
efficiency savings. The speedups observed are less than the
efficiency savings as a result of the computational cost
associated with repeatedly establishing if candidate GP programs
have not \emph{lost} any tournament they are involved in. Clearly
the use of the efficient tournament selection technique has
provided a boost in the performance of GP with a minor speedup in
all cases with a maximum of 15\% achieved.

\section{Consideration of Previously Evaluated
Individuals}\vspace{-0.2cm} \label{sec:nonmod} The previous
section demonstrated that speedups in GP can be achieved using the
efficient tournament selection technique whilst not affecting the
outcome of the GP process but the gains are rather limited as at
least 50\% of the fitness cases need to be evaluated before
\emph{losers} of tournaments can be identified. The fitness metric
is boolean in nature in that a fitness case is either correctly
classified or not so a GP program cannot have
\emph{mathematically} \emph{lost} a tournament whilst 50\% of
fitness cases remain. However, the crossover and mutation
parameters used both had a probability of 0.5. Consequently,
approximately 25\% of candidate GP programs will survive intact
into the next generation and not reevaluating these will in itself
provide an efficiency saving of approximately 25\%. More
importantly, these GP programs are previous tournament
\emph{winners} whereby their complete fitness is known. This will
make it possible to identify \emph{losers} of tournaments at an
earlier stage in the evaluation process. Additionally, as
\emph{winners} of tournaments, they are likely highly fit and
leaving little margin for error for other tournament contenders.
\\\\
\textbf{Example:} Consider a tournament size of two where the
first GP program has survived intact into the next generation and
been previously evaluated correctly classifying eight of the ten
fitness cases. If the second candidate GP program incorrectly
classifies more than two of the fitness cases then it cannot
possibly \emph{win} the tournament. So if the first three fitness
cases are incorrectly classified the solution has \emph{lost} the
tournament and an efficiency saving of 70\% can be achieved.
\\\\
To test this theory the standard approach to GP is rerun as a
benchmark but this time not reevaluating candidate GP programs
which survive intact into the next generation. Additionally,
candidate GP programs which are not involved in any tournament,
the non-sampled issue, are also not evaluated. The results are
shown in Table \ref{tab:NonMod} with an expected 25\% reduction in
execution speed through this efficiency for standard GP. The
comparison results from the efficient tournament selection method
are also shown in Table \ref{tab:NonMod}. Efficiency savings are
now considerably higher than those achieved in Table
\ref{tab:firstResults} even taking into account the efficiency
savings achieved by not reevaluating intact candidate GP programs.
Indeed, the additional efficiency in now as much as 30\%.
Subsequently, it can be considered that having previously
evaluated candidate GP programs within tournaments makes it easier
to establish \emph{losers} of tournaments hence increasing the
efficiency savings. Furthermore, it can also be observed that in
cases of higher selection pressure the efficiency savings increase
further. Indeed, in the case of the KDDcup classification problem,
efficiency savings of 60\% are achieved. The reason for this is
that GP programs that survive intact into the next generation have
won greatly competitive tournaments and are thus highly fit making
it easier to establish \emph{losers} of tournaments. Note that
there is still no change in the evolutionary path when using
efficient tournament selection.

\begin{center}
\begin{table*}[!h]
\scriptsize \centering \caption{Results from a comparison between
a standard implementation of GP and the efficient tournament
selection method taking into account not reevaluating non-modified
or evaluating non-selected candidate GP programs.}
\begin{tabular}{c
                c
                S[table-number-alignment=center,separate-uncertainty,table-figures-uncertainty=1, table-figures-integer = 2, table-figures-decimal = 2]
                c
                S[table-number-alignment=center,separate-uncertainty,table-figures-uncertainty=1, table-figures-integer = 3, table-figures-decimal = 2, table-column-width=20mm]
                S[table-number-alignment=center,separate-uncertainty,table-figures-uncertainty=1, table-figures-integer = 2, table-figures-decimal = 2, table-column-width=19mm]
                S[table-number-alignment=center,separate-uncertainty,table-figures-uncertainty=1, table-figures-integer = 2, table-figures-decimal = 2, table-column-width=19mm]
                c
                }
\toprule {\multirow{3}{1.1cm}{\centering Problem}} &
{\multirow{3}{0.9cm}{\centering Tourn. Size}} &
{\multirow{3}{1.4cm}{\centering Class Accuracy (\%)}} &
{\multirow{3}{1.15cm}{\centering Av. Tree Size}} &
{\multirow{3}{1.7cm}{\centering Standard GP Execution Time (s)}} & \multicolumn{3}{c}{Efficient Tournament Selection GP}\\
\cmidrule{6-8} & & & & & {\multirow{2}{1.6cm}{\centering Efficiency Saving (\%)}} & {\multirow{2}{1.6cm}{\centering Execution Time (s)}} & {\multirow{2}{1.0cm}{\centering Speedup}}\\
& & & & & & &\\
\midrule
\multirow{10}{1.1cm}{Shuttle} & 3 & 83.96 \pm 3.67 & 43.02 & 10.12 \pm 2.08 & 35.73 \pm 1.77 & 9.45 \pm 1.86 & 1.070x \\
& 4 & 87.92 \pm 5.34 & 46.57 & 10.84 \pm 3.26 & 35.38 \pm 2.50 & 10.00 \pm 2.77 & 1.084x \\
& 5 & 88.10 \pm 6.47 & 42.07 & 9.97 \pm 1.62 & 35.92 \pm 2.73 & 9.15 \pm 1.23 & 1.090x \\
& 6 & 89.23 \pm 6.19 & 44.49 & 10.68 \pm 2.62 & 37.37 \pm 2.57 & 9.59 \pm 2.27 & 1.114x \\
& 7 & 88.93 \pm 6.42 & 43.64 & 10.37 \pm 3.19 & 37.24 \pm 4.21 & 9.33 \pm 2.57 & 1.112x \\
& 8 & 91.44 \pm 5.99 & 41.34 & 10.00 \pm 2.31 & 38.99 \pm 4.10 & 8.80 \pm 1.80 & 1.137x \\
& 9 & 92.02 \pm 6.68 & 44.56 & 10.64 \pm 1.88 & 41.23 \pm 4.17 & 9.14 \pm 1.73 & 1.164x \\
& 10 & 90.53 \pm 6.24 & 41.94 & 10.15 \pm 2.70 & 40.42 \pm 4.68 & 8.78 \pm 2.14 & 1.156x \\
& 20 & 90.10 \pm 6.75 & 47.20 & 11.40 \pm 4.18 & 46.56 \pm 8.49 & 9.11 \pm 3.28 & 1.252x \\
& 30 & 86.43 \pm 7.36 & 47.72 & 11.48 \pm 4.19 & 45.03 \pm 8.94 & 9.51 \pm 3.44 & 1.207x \\

\midrule
\multirow{10}{1.1cm}{KDDcup} & 3 & 93.71 \pm 6.84 & 52.03 & 94.84 \pm 20.17 & 38.42 \pm 0.90 & 83.98 \pm 17.62 & 1.129x \\
& 4 & 95.96 \pm 5.37 & 54.03 & 102.27 \pm 20.34 & 38.90 \pm 1.35 & 87.08 \pm 16.51 & 1.174x \\
& 5 & 95.52 \pm 5.29 & 55.04 & 104.58 \pm 26.29 & 39.69 \pm 2.15 & 87.44 \pm 22.24 & 1.196x \\
& 6 & 95.58 \pm 5.78 & 47.65 & 91.98 \pm 18.90 & 41.24 \pm 2.38 & 75.23 \pm 15.59 & 1.223x \\
& 7 & 92.71 \pm 8.09 & 50.78 & 97.57 \pm 24.97 & 41.80 \pm 3.05 & 78.99 \pm 20.32 & 1.235x \\
& 8 & 96.26 \pm 5.24 & 51.91 & 98.74 \pm 27.43 & 44.01 \pm 2.00 & 76.99 \pm 20.86 & 1.283x \\
& 9 & 96.19 \pm 5.36 & 56.98 & 109.29 \pm 37.46 & 45.67 \pm 3.80 & 83.15 \pm 28.72 & 1.314x \\
& 10 & 95.40 \pm 6.28 & 58.16 & 113.60 \pm 47.04 & 48.06 \pm 2.92 & 82.96 \pm 33.30 & 1.369x \\
& 20 & 95.91 \pm 5.37 & 64.90 & 123.26 \pm 64.26 & 58.76 \pm 5.72 & 73.39 \pm 37.33 & 1.680x \\
& 30 & 94.64 \pm 7.12 & 65.66 & 123.74 \pm 59.28 & 57.98 \pm 7.10 & 76.15 \pm 36.99 & 1.625x \\

\midrule
\multirow{10}{1.1cm}{20-Mult.} & 3 & 61.46 \pm 0.88 & 185.11 & 38.17 \pm 5.43 & 31.49 \pm 0.17 & 37.73 \pm 5.33 & 1.012x \\
& 4 & 63.44 \pm 1.18 & 159.25 & 35.14 \pm 5.98 & 29.73 \pm 0.20 & 34.60 \pm 5.85 & 1.016x \\
& 5 & 64.85 \pm 1.35 & 154.58 & 35.19 \pm 5.05 & 29.45 \pm 0.31 & 34.43 \pm 4.90 & 1.022x \\
& 6 & 66.61 \pm 1.78 & 156.98 & 35.63 \pm 6.24 & 29.66 \pm 0.33 & 34.61 \pm 6.02 & 1.029x \\
& 7 & 67.02 \pm 1.75 & 141.64 & 32.58 \pm 4.95 & 30.00 \pm 0.49 & 31.46 \pm 4.80 & 1.036x \\
& 8 & 67.36 \pm 2.06 & 159.75 & 35.77 \pm 9.81 & 30.13 \pm 0.59 & 34.46 \pm 9.58 & 1.038x \\
& 9 & 68.38 \pm 1.74 & 132.88 & 31.54 \pm 5.23 & 30.69 \pm 0.54 & 30.19 \pm 5.10 & 1.045x \\
& 10 & 68.61 \pm 2.25 & 128.93 & 30.73 \pm 6.01 & 31.04 \pm 0.67 & 29.19 \pm 5.78 & 1.053x \\
& 20 & 71.17 \pm 2.06 & 135.81 & 31.64 \pm 6.55 & 34.28 \pm 1.03 & 28.84 \pm 6.21 & 1.097x \\
& 30 & 72.61 \pm 2.22 & 157.60 & 34.71 \pm 7.56 & 35.42 \pm 1.54 & 31.11 \pm 6.80 & 1.116x \\
\bottomrule
\end{tabular} \centering
\label{tab:NonMod}
\end{table*}
\end{center}

\vspace{-1.0cm}

In terms of the effective speed of GP from using these tournament
selection efficiency savings, the greatest increase in speed has
been achieved for the KDDcup problem with a 1.68x performance gain
when using larger tournament sizes. Indeed, for all problem
instances, the best performance gains are observed from higher
selection pressure. The higher the selection pressure, the greater
probability that highly fit candidate GP programs survive intact
into the next generation.

\vspace{-0.4cm}
\subsection{Elitism}
\label{sec:Elite} Given that having candidate GP programs that
have survived intact into subsequent generations has been shown to
make it easier to correctly identify \emph{losers} of tournaments
it could be considered that using the elitism operator would have
a similar effect. Elitism involves the best subset of candidate GP
programs in a given population being copied intact into the next
generation meaning the best solutions are never lost from the
population. As previously, these \emph{elitist} candidate GP
programs do not need to be reevaluated thus providing a basic
efficiency saving. Indeed, Table \ref{tab:NonModElite} shows the
results generated from standard GP using an elitism operator of
10\% of the population and also not evaluating non-modified or
non-sampled candidate GP programs. From these results it can be
seen that an expected approximate 32\% efficiency saving is now
achieved in cases of high selection pressure whereby non-sampling
is not an issue. It should be noted that the use of the elitism
operator has had an effect on the classification accuracy achieved
leading to slight differences when compared to the results in
Table \ref{tab:firstResults}.

\begin{center}
\begin{table*}[!h]
\scriptsize \centering \caption{Results from using standard GP
with 10\% elitism and not reevaluating non-modified or evaluating
non-selected individuals\vspace{-0.2cm}}
\begin{tabular}{c
                c
                S[table-number-alignment=center,separate-uncertainty,table-figures-uncertainty=1, table-figures-integer = 2, table-figures-decimal = 2]
                c
                S[table-number-alignment=center,separate-uncertainty,table-figures-uncertainty=1, table-figures-integer = 3, table-figures-decimal = 2, table-column-width=20mm]
                c
                S[table-number-alignment=center,separate-uncertainty,table-figures-uncertainty=1, table-figures-integer = 2, table-figures-decimal = 2, table-column-width=20mm]
                }
\toprule
{\multirow{2}{1.2cm}{\centering Problem}} & {\multirow{2}{1.0cm}{\centering Tourn. Size}} & {\multirow{2}{2.0cm}{\centering Classification Accuracy (\%)}} & {\multirow{2}{1.25cm}{\centering Av. Tree Size}} & {\multirow{2}{1.75cm}{\centering Execution Time (s)}} & {\multirow{2}{1.1cm}{\centering GPop/s (bn)}} & {\multirow{2}{1.75cm}{\centering Efficiency Saving (\%)}}\\
& & & & &\\
\midrule
\multirow{10}{1.2cm}{Shuttle} & 3 & 87.60 \pm 5.81 & 30.64 & 7.10 \pm 1.82 & 50.88 & 36.48 \pm 0.63 \\
& 4 & 89.86 \pm 5.30 & 33.71 & 7.86 \pm 2.15 & 50.08 & 34.61 \pm 0.61 \\
& 5 & 89.09 \pm 5.07 & 32.71 & 7.78 \pm 1.75 & 49.26 & 33.76 \pm 0.52 \\
& 6 & 89.40 \pm 6.65 & 36.38 & 8.38 \pm 2.28 & 50.36 & 33.61 \pm 0.53 \\
& 7 & 91.73 \pm 4.64 & 37.36 & 8.64 \pm 2.37 & 50.36 & 33.33 \pm 0.57 \\
& 8 & 86.90 \pm 6.84 & 36.01 & 8.33 \pm 2.51 & 50.24 & 33.18 \pm 0.59 \\
& 9 & 88.89 \pm 6.81 & 32.91 & 7.68 \pm 1.97 & 50.05 & 33.05 \pm 0.82 \\
& 10 & 89.20 \pm 6.74 & 38.49 & 8.73 \pm 2.76 & 50.90 & 33.17 \pm 0.78 \\
& 20 & 89.70 \pm 7.11 & 38.46 & 8.98 \pm 2.57 & 49.82 & 33.08 \pm 0.80 \\
& 30 & 90.31 \pm 6.94 & 38.75 & 9.01 \pm 2.68 & 49.02 & 33.07 \pm 0.68 \\

\midrule
\multirow{10}{1.2cm}{KDDcup} & 3 & 93.79 \pm 7.67 & 28.98 & 53.70 \pm 12.28 & 53.59 & 35.68 \pm 0.71 \\
& 4 & 94.76 \pm 7.01 & 38.07 & 69.67 \pm 21.00 & 53.85 & 34.09 \pm 0.35 \\
& 5 & 96.45 \pm 4.12 & 38.15 & 68.80 \pm 21.40 & 54.86 & 33.33 \pm 0.20 \\
& 6 & 92.25 \pm 8.72 & 36.75 & 67.17 \pm 16.15 & 54.28 & 32.73 \pm 0.64 \\
& 7 & 94.39 \pm 6.93 & 39.47 & 73.35 \pm 32.31 & 53.20 & 32.94 \pm 0.38 \\
& 8 & 95.40 \pm 5.30 & 42.71 & 77.21 \pm 30.06 & 54.15 & 32.81 \pm 0.46 \\
& 9 & 94.43 \pm 7.12 & 42.64 & 75.39 \pm 26.44 & 55.95 & 32.67 \pm 0.96 \\
& 10 & 96.47 \pm 4.05 & 43.20 & 76.85 \pm 25.67 & 55.64 & 32.86 \pm 0.34 \\
& 20 & 96.20 \pm 5.33 & 51.49 & 92.24 \pm 34.83 & 55.21 & 33.08 \pm 0.37 \\
& 30 & 95.00 \pm 6.15 & 50.33 & 92.05 \pm 50.41 & 54.57 & 32.97 \pm 0.60 \\

\midrule
\multirow{10}{1.2cm}{20-Mult.} & 3 & 65.22 \pm 1.17 & 117.25 & 26.07 \pm 3.44 & 958.04 & 36.43 \pm 0.34 \\
& 4 & 66.75 \pm 1.24 & 128.45 & 27.86 \pm 5.29 & 973.35 & 34.44 \pm 0.33 \\
& 5 & 67.86 \pm 1.74 & 121.00 & 26.95 \pm 4.93 & 951.26 & 33.71 \pm 0.29 \\
& 6 & 67.87 \pm 1.73 & 113.12 & 25.95 \pm 4.06 & 923.90 & 33.41 \pm 0.34 \\
& 7 & 69.31 \pm 1.48 & 113.86 & 26.25 \pm 3.71 & 923.68 & 33.44 \pm 0.19 \\
& 8 & 69.45 \pm 1.67 & 115.16 & 26.40 \pm 5.50 & 914.43 & 33.33 \pm 0.38 \\
& 9 & 69.73 \pm 2.01 & 125.68 & 27.67 \pm 6.01 & 953.56  & 33.50 \pm 0.26 \\
& 10 & 70.40 \pm 1.57 & 128.21 & 27.52 \pm 6.89 & 974.39 & 33.54 \pm 0.16 \\
& 20 & 73.09 \pm 1.97 & 126.28 & 27.30 \pm 5.09 & 974.77 & 33.59 \pm 0.19 \\
& 30 & 73.80 \pm 2.70 & 140.39 & 29.78 \pm 6.34 & 990.59 & 33.66 \pm 0.19 \\
\bottomrule
\end{tabular} \centering
\label{tab:NonModElite}
\end{table*}
\end{center}

\vspace{-1.0cm}

\begin{center}
\begin{table*}[!h]
\scriptsize \centering \caption{Results from using efficient
tournament selection with 10\% elitism and not reevaluating
non-modified or evaluating non-selected
individuals\vspace{-0.2cm}}
\begin{tabular}{c
                c
                S[table-number-alignment=center,separate-uncertainty,table-figures-uncertainty=1, table-figures-integer = 2, table-figures-decimal = 2]
                c
                S[table-number-alignment=center,separate-uncertainty,table-figures-uncertainty=1, table-figures-integer = 3, table-figures-decimal = 2, table-column-width=20mm]
                c
                S[table-number-alignment=center,separate-uncertainty,table-figures-uncertainty=1, table-figures-integer = 2, table-figures-decimal = 2, table-column-width=20mm]
                c
                }
\toprule
{\multirow{2}{1.2cm}{\centering Problem}} & {\multirow{2}{1.0cm}{\centering Tourn. Size}} & {\multirow{2}{2.0cm}{\centering Classification Accuracy (\%)}} & {\multirow{2}{1.1cm}{\centering Av. Tree Size}} & {\multirow{2}{1.6cm}{\centering Execution Time (s)}} & {\multirow{2}{1.1cm}{\centering GPop/s (bn)}}  & {\multirow{2}{1.6cm}{\centering Efficiency Saving (\%)}}& {\multirow{2}{1.0cm}{\centering Speedup}}\\
& & & & & & & \\
\midrule
\multirow{10}{1.2cm}{Shuttle} & 3 & 87.17 \pm 5.14 & 30.55 & 6.60 \pm 1.07 & 54.68 & 44.26 \pm 2.21 & 1.076x \\
& 4 & 90.09 \pm 5.06 & 33.53 & 7.14 \pm 1.78 & 54.82 & 44.32 \pm 2.51 & 1.100x \\
& 5 & 88.77 \pm 4.85 & 32.50 & 6.89 \pm 1.72 & 54.94 & 45.04 \pm 2.74 & 1.129x \\
& 6 & 88.96 \pm 6.76 & 36.90 & 7.40 \pm 2.14 & 57.43 & 46.08 \pm 3.72 & 1.133x \\
& 7 & 91.36 \pm 4.24 & 35.23 & 6.95 \pm 1.45 & 58.96 & 48.59 \pm 4.08 & 1.244x \\
& 8 & 87.47 \pm 7.09 & 36.72 & 7.33 \pm 2.15 & 58.26 & 46.83 \pm 4.49 & 1.135x \\
& 9 & 88.94 \pm 7.06 & 32.03 & 6.46 \pm 1.52 & 57.57 & 48.55 \pm 6.13 & 1.187x \\
& 10 & 89.21 \pm 6.75 & 38.25 & 7.25 \pm 2.26 & 61.17 & 49.77 \pm 4.61 & 1.204x \\
& 20 & 89.70 \pm 7.11 & 38.46 & 7.14 \pm 2.16 & 63.24 & 54.26 \pm 8.86 & 1.257x \\
& 30 & 90.31 \pm 6.94 & 38.75 & 7.23 \pm 1.95 & 61.20 & 53.11 \pm 7.67 & 1.247x \\

\midrule
\multirow{10}{1.2cm}{KDDcup} & 3 & 95.41 \pm 6.31 & 33.42 & 50.71 \pm 12.55 & 64.41 & 47.69 \pm 1.58 & 1.059x \\
& 4 & 95.30 \pm 6.03 & 32.17 & 48.72 \pm 8.02 & 65.89 & 49.20 \pm 1.69 & 1.430x \\
& 5 & 94.37 \pm 7.02 & 38.41 & 54.49 \pm 18.42 & 69.87 & 50.38 \pm 2.60 & 1.263x \\
& 6 & 94.42 \pm 7.01 & 38.48 & 53.65 \pm 12.75 & 71.95 & 52.26 \pm 3.13 & 1.252x \\
& 7 & 95.75 \pm 5.34 & 41.19 & 55.36 \pm 17.52 & 74.18 & 54.09 \pm 3.69 & 1.325x \\
& 8 & 95.94 \pm 5.20 & 39.11 & 50.52 \pm 13.79 & 76.67 & 56.07 \pm 3.15 & 1.528x \\
& 9 & 95.63 \pm 6.30 & 40.34 & 50.38 \pm 20.19 & 80.33 & 58.17 \pm 4.74 & 1.496x \\
& 10 & 96.14 \pm 5.36 & 41.67 & 50.64 \pm 17.67 & 81.53 & 59.24 \pm 5.23 & 1.518x \\
& 20 & 96.20 \pm 5.33 & 51.49 & 52.94 \pm 18.57 & 95.86 & 65.50 \pm 4.28 & 1.742x \\
& 30 & 95.00 \pm 6.15 & 50.33 & 54.58 \pm 27.23 & 91.24 & 64.68 \pm 5.89 & 1.686x \\

\midrule
\multirow{10}{1.2cm}{20-Mult.} & 3 & 65.29 \pm 1.15 & 118.13 & 25.52 \pm 4.35 & 977.80 & 38.45 \pm 0.42 & 1.021x \\
& 4 & 66.55 \pm 1.43 & 129.45 & 27.37 \pm 5.50 & 997.36 & 37.09 \pm 0.36 & 1.018x \\
& 5 & 68.11 \pm 1.79 & 122.27 & 25.98 \pm 4.22 & 994.67 & 36.98 \pm 0.29 & 1.037x \\
& 6 & 67.87 \pm 1.59 & 115.87 & 24.97 \pm 3.46 & 980.32 & 37.15 \pm 0.62 & 1.039x \\
& 7 & 69.44 \pm 1.59 & 117.88 & 25.29 \pm 4.40 & 984.98 & 37.81 \pm 0.60 & 1.038x \\
& 8 & 69.67 \pm 1.69 & 115.97 & 25.00 \pm 5.24 & 972.22 & 38.24 \pm 0.75 & 1.056x \\
& 9 & 69.82 \pm 1.86 & 123.76 & 25.78 \pm 5.88 & 1007.83 & 38.56 \pm 0.66 & 1.073x \\
& 10 & 70.43 \pm 1.57 & 127.35 & 25.48 \pm 6.42 & 1048.23 & 39.25 \pm 0.79 & 1.080x \\
& 20 & 73.09 \pm 1.97 & 126.28 & 24.44 \pm 4.74 & 1089.81 & 41.82 \pm 1.31 & 1.117x \\
& 30 & 73.80 \pm 2.70 & 140.39 & 26.45 \pm 5.71 & 1115.85 & 42.50 \pm 1.60 & 1.126x \\
\bottomrule
\end{tabular} \centering
\label{tab:NonModEliteMathematical}
\end{table*}
\end{center}

\vspace{-1.2cm}

It should be expected from the earlier results that the elitism
operator will improve the savings of the efficient tournament
selection model further and as such the experiments are rerun
using 10\% elitism with the results shown in Table
\ref{tab:NonModEliteMathematical}. It should be firstly noted that
the classification accuracy and average tree sizes differ slightly
to that those of the standard GP approach as shown in Table
\ref{tab:NonModElite}, a divergence in the evolutionary process
has occurred. The reason behind this is that a highly fit
candidate GP program capable of being selected by the elitism
operator was not selected because it \emph{lost} all the
tournaments it was involved in and hence its continued evaluation
terminated resulting in a lower fitness level and thereby no
longer qualifying for selection by elitism. Normally, it would be
selected even though it would not \emph{win} any tournaments.
However, the effect on classifier accuracy is minimal and in some
cases the accuracy is actually improved. Efficiency savings of
40-65\% are achieved for all problem instances with a peak
occurring for the KDDcup classification problem and a high level
of selection pressure whereby highly fit solutions are more
influential. Indeed, for all problem instances, the greatest
efficiency savings are once again achieved for high selection
pressure. It should be noted though that when using increased
tournament sizes, the average execution time tends to be greater
as a result of larger GP trees being considered. Thus, a tradeoff
needs to be considered between efficiency savings and the increase
in the size of the average GP tree.

From the results observed in Table \ref{tab:NonModElite}, using
the elitism operator should provide an additional efficiency of
approximately 7\%. Observing the efficiency savings in Table
\ref{tab:NonModEliteMathematical} and comparing with the previous
results in Table \ref{tab:NonMod}, efficiency savings have
improved from between 8\% and 12\%. Thereby, it can be considered
that use of the elitism operator has further benefited
identification of candidate GP programs who \emph{mathematically}
cannot \emph{win} tournaments at an earlier point in the
evaluation phase. In terms of performance compared to a standard
implementation of GP which uses elitism and does not reevaluate
non-modified candidate GP programs, classifier training on the
KDDcup problem occurs up to 1.74x faster. A brief mention of
Genetic Programming Operations per Second (GPop/s) should be made
with a maximum effective GP speed of 96 billion GPop/s achieved
for the KDDcup classification problem and 1116 billion GPop/s for
the multiplexer problem benefiting from an extra 32x bitwise
parallelism.

\vspace{-0.2cm}\section{Conclusions}\vspace{-0.2cm} In this paper
a methodology for evaluating candidate GP programs is presented
which provides significant computational efficiency savings even
when embedded within a high performance parallel model. This
methodology exploits tournament selection in that it is possible
to identify \emph{losers} of tournaments before evaluation of all
the fitness cases is complete. Essentially, by evaluating all GP
programs on subsets of fitness cases before moving to the next
subset, comparisons can be made between solutions and hence
\emph{losers} of tournaments can be identified and early
termination of the evaluation of these solutions achieved. This
approach was shown to provide minor efficiency savings and hence
runtime speedups. However, this paper discovers that the true
advantage of the technique arises when solutions survive intact
into the next generation. These solutions have \emph{won}
tournaments so are highly fit with a known fitness enabling much
earlier detection of \emph{losers} of tournaments especially when
using high selection pressure. Efficiency savings of up to 65\%
and subsequent speedups in execution speed of up to 1.74x were
demonstrated with a peak rate of 96 billion GPop/s achieved by GP
running on a multi-core CPU. Further work needs to consider
alternative fitness metrics such as the Mean Squared Error (MSE)
for regression problems, combining the technique with sampling
approaches and alternative methods for correctly predicting
tournament \emph{losers} earlier.

\vspace{-0.2cm}\section{Acknowledgement}\vspace{-0.2cm} This is a
pre-print of a contribution published in Lotfi A., Bouchachia H.,
Gegov A., Langensiepen C., McGinnity M. (eds) Advances in
Computational Intelligence Systems. UKCI 2018. Advances in
Intelligent Systems and Computing, vol 840 published by Springer.
The definitive authenticated version is available online via
https://doi.org/10.1007/978-3-319-97982-3\_4.

\vspace{-0.2cm}

%\begin{thebibliography}{6}
%
\scriptsize
\bibliographystyle{splncs03}
\bibliography{ExploitingTournamentSelection}
\vspace{-0.3cm}
%\end{thebibliography}
\end{document}